%% file: ecai-sample-and-instructions.tex
\documentclass{ecai} 
\usepackage{microtype}
\usepackage{graphicx}
\usepackage{booktabs} 
\usepackage{multirow} 
\usepackage{subcaption} 
\usepackage{mathtools}
\usepackage{float}
\usepackage{bm}

\usepackage{latexsym}
\usepackage{amssymb}
\usepackage{amsmath}
\usepackage{amsthm}
\usepackage{booktabs}
\usepackage{enumitem}
\usepackage{graphicx}
\usepackage{color}
\usepackage{hyperref}
\usepackage{algorithm}
\usepackage{algpseudocode}

\usepackage[T1]{fontenc}
\usepackage{graphicx}
\usepackage{booktabs}
\usepackage[misc]{ifsym}

\input{math_commands.tex}

\usepackage{hyperref}
\usepackage{url}
\usepackage{enumitem}

\usepackage{algorithm}
\usepackage{algpseudocode} 

\usepackage{caption}  
\usepackage{multirow} 
\usepackage{amssymb}

\usepackage{amsmath}  
\usepackage{booktabs} 
\usepackage{array}    

\newcommand{\BibTeX}{B\kern-.05em{\sc i\kern-.025em b}\kern-.08em\TeX}

\begin{document}


\begin{frontmatter}


\paperid{1} 


\title{eMargin: Revisiting Contrastive Learning with Margin-Based Separation}


\author[A]{\fnms{Abdul-Kazeem}~\snm{Shamba}\thanks{Corresponding Author. Email: abdul.k.shamba@ntnu.no.}}
\author[A]{\fnms{Kerstin}~\snm{Bach}}
\author[B]{\fnms{Gavin}~\snm{Taylor}}

\address[A]{Department of Computer Science, Norwegian University of Science and Technology, Norway}
\address[B]{Department of Computer Science, United States Naval Academy, USA}


\begin{abstract}
We revisit previous contrastive learning frameworks to investigate the effect of introducing an adaptive margin into the contrastive loss function for time series representation learning. Specifically, we explore whether an adaptive margin (\textit{eMargin}), adjusted based on a predefined similarity threshold, can improve the separation between adjacent but dissimilar time steps and subsequently lead to better performance in downstream tasks. Our study evaluates the impact of this modification on clustering performance and classification in three benchmark datasets. Our findings, however, indicate that achieving high scores on unsupervised clustering metrics does not necessarily imply that the learned embeddings are meaningful or effective in downstream tasks. To be specific, \textit{eMargin} added to InfoNCE consistently outperforms state-of-the-art baselines in unsupervised clustering metrics, but struggles to achieve competitive results in downstream classification with linear probing. The source code is publicly available at \href{https://github.com/sfi-norwai/eMargin}{https://github.com/sfi-norwai/eMargin}.

\end{abstract}

\end{frontmatter}


\section{Introduction}

Self-supervised learning (SSL) takes advantage of the inherent structure of unannotated data to produce rich feature embeddings \cite{bengio,npair,byol,dino,jepa}. Contrastive learning (CL) \cite{hadsell2006} is one of the SSL techniques. The major principle in CL is to learn a representation space where similar (positive) samples are pulled closer together while dissimilar (negative) samples are pushed apart. This is typically done using contrastive loss, such as InfoNCE \cite{cpc}. For computer vision tasks, contrastive learning is highly dependent on data enhancement to generate positive pairs. For example, SimCLR \cite{simclr} uses random crops, rotations, and color distortions to create different views of the same image,assuming that they represent the same underlying concept.

Unlike images where augmentations like flipping or cropping do not necessarily alter the semantic meaning, time series data are highly structure- and order-dependent. Many works have exploited these temporal correlations to learn meaningful representations in time series \cite{tnc,cpc,ts2vec,infots}. However, naively contrasting adjacent timestamps or similar instances within a time series ignores their temporal dependencies, leading to suboptimal representations.

Although it is possible to introduce margin in the contrastive objective, recent works have sought to address this problem by introducing soft assignments \cite{soft}. However, if positive pairs sampled from immediate pairs are treated equally even during transition with consecutive steps exhibiting high dissimilarities, the model may learn trivial representations that overfit to local continuity rather than capture high-level structures. In this work, we investigate the introduction of margin at transition points in contrastive learning. We refer to this as \textit{eMargin} (adaptive margin) to distinguish it from the static margin used in previous works. While eMargin encourages increased inter-class variance and reduced intra-class variance, leading to more compact and well-separated clusters, our findings reveal a critical disconnect between clustering quality and downstream task performance. Specifically, although eMargin improves unsupervised clustering metrics, it often results in poor performance on downstream classification. Our contributions are as follows.

\begin{itemize}[left=0pt]

   \item We incorporate an adaptive margin into the contrastive learning objective and evaluate its impact on learned embeddings.

    \item We conduct experiments on three real-world time series datasets and show that while eMargin significantly improves unsupervised clustering metrics such as DBI and Silhouette score, it consistently underperforms in downstream classification using linear probing—highlighting a misalignment between clustering-based evaluation and downstream utility.

\end{itemize}

\section{Related Works}
\label{sec4}

\paragraph{\textbf{Margin in Contrastive Learning}} The use of margins to regulate feature separation has been explored in both metric learning and contrastive frameworks. Traditional metric learning methods, such as triplet loss \cite{Schroff_2015_CVPR}, enforce a fixed margin between positive and negative pairs to improve discriminability. While effective in supervised settings, fixed margins struggle with self-supervised scenarios where similarity structures are dynamic and task-agnostic. Recent contrastive learning works have incorporated margins to address representation collapse. For instance, \cite{robinson2021contrastivelearninghardnegative} proposed a margin-based contrastive loss to penalize hard negatives, ensuring stable training by preventing over-suppression of semantically similar pairs. Similarly, \cite{khosla2021supervisedcontrastivelearning} extended contrastive learning to supervised settings with class-aware margins, although their approach assumes label availability and static margin design.  More recent approaches, such as \cite{soft}, address this by incorporating a soft contrastive loss that dynamically scales similarity penalties based on time-dependent correlations. This work builds upon these efforts by investigating the effect of adaptive margin for temporal contrastive learning, dynamically adjusting feature separations based on similarity within a sequence. 

\paragraph{\textbf{Contrastive learning in time series}} With the recent traction of CL in CV and NLP, several works in TS representation learning have proposed different methods for sampling positive and negative pairs. \cite{mixup} creates a new augmented sample of a time series and attempts to predict the strength of the mixing components. \cite{tfc} samples positive pairs as time-based and frequency-based representations from the time series signal and introduces a time-frequency consistency framework. \cite{timeclr} introduces dynamic time warping (DTW) data augmentation for creating phase shifts and amplitude changes. To learn discriminative representation across time, TS2Vec \cite{ts2vec} considers the representation at the same time stamp from two views as positive pairs. InfoTS \cite{infots} focuses on developing criteria for selecting good augmentation in contrastive learning in the TS domain. T-loss \cite{tloss} employs a time-based sample and a triplet loss to learn representation by selecting positive and negative samples based on their temporal distance from the anchor. TNC \cite{tnc} presents temporal neighborhood with a statistical test to determine the neighborhood range that it treats as positive samples. \cite{ncl}, on the other hand, selects neighbors based on both instance-level and temporal-level criteria with a trade-off parameter allowing the model to balance instance-wise distinction with temporal coherence. \cite{clocs} define a positive pair as a representation of transformed instances of the same subject. TS-TCC \cite{tstcc} proposes a method to combine temporal and contextual information in TS using data augmentation to select positives and predict the future of one augmentation using past features of another representation in the temporal contrasting module. CoST \cite{cost} applied CL in learning representation for TS forecasting by having inductive biases in model architecture to learn disentangled seasonal trends.

\section{Margin in Contrastive Learning}
\label{headings}

We aim to understand the role of adaptive margin in contrastive learning. Given an input sequence of length \(T\), where consecutive time steps are treated as positives, contrastive learning seeks to learn a representation function \( f(.) \) that maps these time steps into an embedding space, such that \( z_t = f(x_t) \). The extended InfoNCE loss as in \cite{simclr} is given by:

\begin{equation}
    L_{\text{InfoNCE}} = -\mathbb{E} \left[ \log \frac{\exp(\text{sim}(z_t, z_{t+1}) / \tau)}{\sum_{k \neq t, t+1} \exp(\text{sim}(z_t, z_k) / \tau)} \right]
    \label{eqn:infonce}
\end{equation}

where:

\begin{itemize}
    \item \( z_t \) and \( z_{t+1} \) are positive pairs (e.g., adjacent time steps),
    \item \( z_k \) are negative samples,
    \item \( \text{sim}(z_t, z_{t+1}) \) is the cosine similarity,
    \item \( \tau \) is the temperature parameter,
\end{itemize}

This formulation, however, treats all positive pairs equally, leading to low separability even during transition, where consecutive steps exhibit high dissimilarities, leading to trivial representations that exploit local continuity rather than capturing meaningful high-level structure. To address this, we introduce an adaptive margin adjustment that separates features in representation space when adjacent time steps exhibit some degree of dissimilarity in the data space.




\subsection{Defining the Pseudo-Label for Dynamic Margins}

Given a time series input $x$ of length $T$, we define the pseudo-label based on the raw input data rather than the feature space. The pseudo-label $\mathcal{Y}$ is computed using the similarity between consecutive time steps in data-space:

\[
\text{sim}(x_t, x_{t+1}) = \frac{x_t \cdot x_{t+1}}{\|x_t\| \|x_{t+1}\|}
\]

where $x_t, x_{t+1} \in \mathbb{R}^d$ are two consecutive raw data vectors.

To differentiate between highly similar and less similar adjacent time steps, we introduce a pseudo-label $\mathcal{Y}$:

\[
\mathcal{Y} = 
\begin{cases}
0, & \text{if } \text{sim}(x_t, x_{t+1}) > \text{threshold} \\
1, & \text{otherwise}
\end{cases}
\]
where the threshold is a hyperparameter controlling sensitivity to similarity. If two adjacent time steps exceed this threshold, they are treated as strongly correlated (no margin adjustment), otherwise, we enforce a margin to increase their separation.

\subsection{Incorporating the Margin into the Similarity Matrix}

We define a similarity matrix $\mM \in \mathbb{R}^{T \times T}$, where element $\mM_{t,t+1}$ stores the similarity score between consecutive time steps:
\[
\mM_{t,t+1} = \text{sim}(z_t, z_{t+1})
\]
To prevent representation collapse while ensuring sufficient separation for dissimilar samples, we modify this matrix using an adaptive margin penalty:

\begin{equation}
\mM_{\text{margin}} = \frac{1}{2}(1 - \mathcal{Y})\mM^2 + \frac{1}{2} \mathcal{Y} \left[\max(0, \text{margin} - \mM)\right]^2
\end{equation}

This equation consists of two terms: 

\begin{itemize}
    \item The first term, $\frac{1}{2} (1 - \mathcal{Y}) \mM^2$, applies when $\mathcal{Y} = 0$, meaning the time steps are highly similar. In this case, the similarity score remains squared to emphasize the connection. 

    \item The second term, $\frac{1}{2} \mathcal{Y} [\max(0, \text{margin} - \mM)]^2$, applies when $\mathcal{Y} = 1$, meaning the time steps are dissimilar. Here, we introduce a \textit{penalty} that forces the similarity score to be at least the specified margin, increasing separation.
    
\end{itemize}

By modifying the similarity matrix $\mM$ dynamically, we effectively reshape the similarity distribution before applying the softmax operation. This alters the optimization landscape, reducing the risk of trivial solutions where the model overly relies on local smoothness.

\subsection{Mathematical Justification}

To ensure stable learning dynamics, we analyze the behavior of $\mM_{\text{margin}}$ in different regimes:

\begin{itemize}
    \item When $\text{sim}(x_t, x_{t+1})$ is high ($\mathcal{Y} = 0$):
    \[
    \mM_{\text{margin}} = \frac{1}{2} \mM^2
    \]
    This maintains the similarity structure for strongly correlated time steps, reinforcing local continuity.
    
    \item When $\text{sim}(x_t, x_{t+1})$ is low ($\mathcal{Y} = 1$):
    \[
    \mM_{\text{margin}} = \frac{1}{2} \left[\max(0, \text{margin} - \mM)\right]^2
    \]
    If $\mM$ is small, the margin enforces a minimum distance, preventing collapsed representations. If $\mM$ is already larger than the margin, no penalty is applied, allowing flexibility.
\end{itemize}

This approach balances structural consistency (preserving correlations where appropriate) and \textit{representation diversity} (ensuring meaningful separation when needed).

\begin{algorithm}
\caption{Dynamic Margin Adjustment in Contrastive Learning}
\begin{algorithmic}[1]
\Require Time series input $x$ of length $T$, encoder function $f(.)$, threshold, margin
\Ensure Adjusted similarity matrix $\mM_{\text{margin}}$
\For{each consecutive pair $(x_t, x_{t+1})$ in sequence}
    \State Compute pseudo-label:
    \[
    \mathcal{Y} \leftarrow 
    \begin{cases} 
    0, & \text{if } \text{sim}(x_t, x_{t+1}) > \text{threshold} \\ 
    1, & \text{otherwise} 
    \end{cases}
    \]
\EndFor
\State Compute feature representations: $z_t = f(x_t)$ for $t \in \{1, \dots, T\}$
\State Compute similarity matrix $\mM$ where $\mM_{t,t+1} = \text{sim}(z_t, z_{t+1})$
\For{each consecutive pair $(z_t, z_{t+1})$ in sequence}

    \State Apply margin adjustment:
    \[
    \mM_{\text{margin}, t,t+1} \leftarrow 
    \begin{cases} 
    \frac{1}{2} \mM_{t,t+1}^2, & \text{if } \mathcal{Y} = 0 \\ 
    \frac{1}{2} \left[\max(0, \text{margin} - \mM_{t,t+1})\right]^2, & \text{if } \mathcal{Y} = 1 
    \end{cases}
    \]

    \State Compute InfoNCE + eMargin:  
    \[
    L_{\text{eMargin}, t, t+1} = -\mathbb{E} \left[ \log \frac{\exp(\mM_{\text{margin}, t,t+1} / \tau)}{\sum_{k \neq t, t+1} \exp(\mM_{\text{margin}, t, k} / \tau)} \right]
    \]
\EndFor
\State Return adjusted similarity matrix $L_{\text{eMargin}}$
\end{algorithmic}
\label{eqn:algo}
\end{algorithm}

\section{Real-World Time Series Representation Learning}
\label{sec:real}

We investigate time series representation learning on three public time series datasets: \textsc{Harth}, \textsc{Ecg}, and \textsc{SleepEeg}. We preprocess these datasets using different window and hop lengths in the STFT. The \textsc{Harth} dataset is processed to have a sequence length of 119 instances (each instance is 1 second, but with half a second overlap during preprocessing via STFT), corresponding to 60 seconds on the original signal, with a feature dimension of 156. The \textsc{Ecg} dataset, on the other hand, has a sequence length of 500 instances, corresponding to 250 seconds, with a feature dimension of 252. Lastly, the \textsc{SleepEeg} dataset has a sequence length of 300 instances, with each instance representing 200 window size of the signal, and a feature dimension of 178. For pretraining, we split the datasets as follows: a 50-50 train-test split for Harth and an 80-20 train-test split for both SleepEeg and ECG. For model evaluation on clustering, we set aside a balanced subset of all three datasets, which were not used during the training of the unsupervised representation learning model. Specifically, for the \textsc{Harth}, we randomly select 1000 instances of all classes, except class 11 (insufficient samples), where we select 500 random samples. Similarly, for the \textsc{SleepEeg}, we select random 1000 instances from all classes. Finally, for the  \textsc{Ecg}, we select random 1\% of both majority classes and the entirety of the remaining classes to give a distribution of 1577, 972, 459, and 1467 for all four classes, respectively. We conduct all training and evaluation in a fully unsupervised manner, without using labels. To obtain pseudolabels, we apply a threshold to the cosine similarity between consecutive time steps, as outlined in Algorithm \ref{eqn:algo}.

\paragraph{\textbf{Network Architecture}} We employ a straightforward convolutional neural network (CNN) as the feature extractor backbone in the encoder \( E_\theta(\cdot) \). The CNN comprises three blocks of 1D convolution with a kernel size of 1, each followed by batch normalization and ReLU activation, resulting in an embedding dimension of 32. Since our primary focus is on evaluating the objective function, we ensure consistency by using the same architecture for all baselines. Prior to encoding, we preprocess the time series signal using a short-time Fourier transform (STFT), producing an input of dimension \( B \times T \times D \), where \( B \) represents the batch size, \( T \) denotes the sequence length, and \( D \) corresponds to the input dimension.

\paragraph{\textbf{Training Details}}  We train all models using an NVIDIA V100 GPU.  Across all models, we maintain consistent hyperparameters to ensure fair comparison. Specifically, we use a batch size of 8 and a learning rate of 0.001. The number of optimization iterations is set to 200 for datasets with fewer than 160,000 samples and 600 otherwise. The representation dimension is fixed at 320, following \cite{ts2vec}. For optimization, we employ the AdamW optimizer with a margin of 5 and a threshold of 0.4. The temperature parameter $\tau$ is set to 0.05 for Harth, 0.1 for SleepEeg, and 0.5 for ECG. To ensure reproducibility, we train all models using three random seeds (1, 2, and 3).

\subsection{Clusterability}

Following the evaluation in \cite{tnc} we check the properties of the distribution of the representations in the encoding space using unsupervised clustering evaluation metrics. \cite{bengio} posit that the formation of natural clustering is one of the properties of a good representation. We compare with eight state-of-the-art baselines in time series representation learning, InfoTS \cite{infots}, SimMTM \cite{simmtm}, TNC \cite{tnc}, TS2Vec \cite{ts2vec}, Soft \cite{soft}, TimeDRL \cite{timedrl}, CoST \cite{cost} and MF-CLR \cite{mfclr}, to investigate the effect of margin in contrastive learning for sequential data. To capture the performance of each baseline on clustering, we use two popular unsupervised clustering evaluation metrics, namely Davies-Bouldin Index (DBI) \cite{dbi} and Silhouette Score (SS) \cite{ss}. DBI measures the average similarity ratio of each cluster with its most similar cluster. A lower DBI score indicates better separation between clusters. SS evaluates how similar an object is to its own cluster compared to other clusters. SS values range from -1 to 1, with higher values reflecting both compactness and separation. Table \ref{tab:cluster} shows the result of adding margin to the InfoNCE objective against the baseline methods on these unsupervised clustering measures.

\begin{table*}[ht]
\centering
\caption{Clustering performance across three datasets (Harth, SleepEeg, and ECG) is evaluated using the Davies-Bouldin Index (DBI, lower is better) and Silhouette score (higher is better).}

\resizebox{0.99\textwidth}{!}{ 
  \begin{tabular}{lccccccc}
    \toprule
    & \multicolumn{2}{c}{\textbf{\textsc{Harth}}} & \multicolumn{2}{c}{\textbf{\textsc{SleepEeg}}} & \multicolumn{2}{c}{\textbf{\textsc{Ecg}}} \\
    \cmidrule(lr){2-3} \cmidrule(lr){4-5} \cmidrule(lr){6-7} 
    & DBI$\downarrow$ & Silhouette $\uparrow$ & DBI$\downarrow$ & Silhouette $\uparrow$& DBI$\downarrow$ & Silhouette$\uparrow$ \\
    \midrule

     TNC & 1.10±0.11 & 0.40±0.08 & 1.18±0.05 & 0.41±0.10 & 1.06±0.13 & 0.44±0.05 \\
     InfoTS & 1.17±0.07 & 0.29±0.00 & 0.85±0.04 & 0.48±0.01 & 0.73±0.18 & 0.56±0.03 \\
    CoST & 1.91±0.12 & 0.24±0.06 & 1.21±0.03 & 0.40±0.01 & 1.61±0.25 & 0.33±0.02 \\
   
    SimMTM & 1.98±0.11 & \textbf{0.75±0.05} & 1.77±0.19 & 0.49±0.05 & 2.42±0.15 & 0.17±0.01 \\
    TimeDRL & 1.46±0.12 & 0.30±0.04 & 0.65±0.03 & 0.62±0.09 & 0.96±0.11 & 0.45±0.05 \\
    
    TS2Vec & 1.17±0.06 & 0.33±0.02 & 1.11±0.05 & 0.33±0.00 & 0.94±0.03 & 0.45±0.02 \\
    SoftCLT      &1.37±0.01  &0.28±0.02  &1.46±0.10  &0.24±0.02 &1.06±0.07 & 0.39±0.02\\

    MF-CLR  & 1.81±0.10 & 0.23±0.02 & 1.10±0.12 & 0.43±0.02 & 1.27±0.14 & 0.40±0.02 \\

    \midrule

    InfoNCE + \textit{eMargin}  & \textbf{0.97±0.07} & 0.38±0.01 & \textbf{0.62±0.05} & \textbf{0.83±0.04} & \textbf{0.59±0.02} & \textbf{0.80±0.01} \\

    \bottomrule
  \end{tabular}
}
\label{tab:cluster}
\end{table*}

The addition of \textit{eMargin} to InfoNCE consistently improves clustering performance across all datasets, as shown in Table~\ref{tab:cluster}. In terms of the Davies-Bouldin Index (DBI), which measures cluster separation (lower is better), \textit{eMargin} achieves the lowest values across all datasets, indicating improved inter-cluster distinction. For example, on the ECG dataset, InfoNCE+\textit{eMargin} reports a DBI of 0.59 and a Silhouette score of 0.80, outperforming all baselines under both metrics. Similar patterns are observed on Harth and SleepEeg. Compared to baselines, while some methods such as SimMTM attain high Silhouette scores (e.g., 0.75 on Harth), they exhibit relatively high DBI, indicating suboptimal cluster separation. In contrast, \textit{eMargin} consistently achieves both lower DBI and higher Silhouette scores, suggesting a well-balanced improvement in clustering quality.

\subsection{Linear evaluation with frozen backbone}

The main goal  of representations learning is to learn representations that are useful in downstream tasks. With that in mind, we train a linear classifier on top of the learned representations to assess how well the learned features generalize to the task of interest when used by a simple classifier.

\begin{table*}[ht]
\centering
\caption{ Comparison with state-of-the-art methods on linear evaluation with a frozen backbone. We compare eMargin with baselines and a randomly initialized encoder (\textit{Random Init.}) on frozen evaluation. We train a linear classifier on top of the features from the encoder using the entire train set and evaluate the test set. }
\vskip 0.15in
\resizebox{0.9\textwidth}{!}{ 

  \tiny 
  \begin{tabular}{llcccc}
    \toprule
    \textbf{Datasets} & \textbf{Models} & \textbf{Accuracy} &  \textbf{F1 score} & \textbf{Precision} & \textbf{Recall}\\
    \midrule

    \multirow{8}{*}{\textbf{\textsc{Harth}}} 
    & Random Init.  &86.69±0.50  &0.84±0.01  &0.55±0.01 &0.45±0.02\\

    & TNC        &89.03±0.03  &0.88±0.00  &0.70±0.02  &0.55±0.01 \\
    
    & InfoTS         &\textbf{90.12±0.13}  &\textbf{0.89±0.00}  &\textbf{0.71±0.02}  &\textbf{0.56±0.01} \\
    & CoST         &88.68±0.41  &0.88±0.00  &0.64±0.01  &0.56±0.00 \\
    & SimMTM    &68.14±4.83  &0.60±0.08  &0.34±0.04  &0.24±0.05 \\
    & TimeDRL     &71.92±2.37  &0.63±0.04  &0.37±0.04  &0.28±0.01 \\
    & TS2Vec     &87.59±0.23  &0.85±0.00  &0.58±0.01  &0.49±0.00 \\

    & TS2Vec + \textit{SoftCLT}     &88.07±0.13  &0.86±0.00  &0.62±0.03  &0.50±0.01 \\
    
    & MF-CLR        &77.07±5.38  &0.73±0.07& 0.52±0.04  &0.37±0.05 \\

    & InfoNCE + \textit{eMargin}        &87.87±0.54  &0.86±0.01&0.56±0.04  &0.48±0.03 \\

    \midrule
    
    \multirow{8}{*}{\textbf{\textsc{SleepEeg}}} 
    & Random Init.  &46.89±11.34  &0.44±0.08  &0.36±0.03  &0.32±0.04 \\

    & TNC        &55.13±0.50  &\textbf{0.54±0.01}  &0.33±0.01  &\textbf{0.36±0.02} \\
    
    & InfoTS         &43.32±9.66  &0.42±0.07  &\textbf{0.41±0.05}  &0.35±0.03 \\
    & CoST         &56.63±2.56  &0.55±0.00  &0.38±0.02  &0.35±0.04 \\
    & SimMTM    &57.08±0.68  &0.50±0.01  &0.31±0.03  &0.25±0.01 \\
    & TimeDRL     &\textbf{57.30±2.22}  &0.48±0.02  &0.36±0.09  &0.24±0.02 \\
    & TS2Vec     &51.18±1.12  &0.50±0.00  &0.29±0.01  &0.35±0.02 \\

    &  TS2Vec + \textit{SoftCLT}     &53.60±2.44  &0.52±0.01  &0.34±0.03  &0.37±0.02 \\

    & MF-CLR        &57.07±1.89  &0.48±0.02&0.27±0.06  &0.27±0.01 \\
    
    & InfoNCE + \textit{eMargin}        &55.46±0.26  &0.45±0.01&0.18±0.03  &0.23±0.00 \\
    
    \midrule

    \multirow{8}{*}{\textbf{\textsc{Ecg}}} 
   & Random Init.  &65.48±1.44  &0.64±0.01  &0.35±0.01  &0.34±0.01  \\
    & TNC        &75.86±2.00  &0.75±0.02  &0.39±0.00  &0.38±0.01 \\
    & InfoTS         &61.94±3.19  &0.59±0.05  &0.34±0.01  &0.32±0.01 \\
    
    & CoST         &69.37±9.55  &0.67±0.13  &0.38±0.02  &0.36±0.04 \\
    & SimMTM    &71.89±2.27  &0.71±0.03  &0.37±0.00  &0.36±0.01 \\
    & TimeDRL     &59.51±6.93  &0.55±0.10  &0.33±0.02  &0.31±0.03 \\
    & TS2Vec     &\textbf{77.72±1.27}  &\textbf{0.77±0.01}  &\textbf{0.40±0.00}  &\textbf{0.39±0.01} \\

    & TS2Vec + \textit{SoftCLT}      &76.37±3.88  &0.76±0.04  &0.39±0.02  &0.39±0.02 \\

    & MF-CLR        & 52.69±2.29  &0.45±0.06&0.31±0.01  &0.28±0.01 \\
    
    & InfoNCE + \textit{eMargin}        &62.41±0.57  &0.61±0.01&0.34±0.01  &0.32±0.00 \\
    
    \bottomrule
  \end{tabular}
    }
\vskip -0.1in
\label{tab:linear}
\end{table*}

The results in Table~\ref{tab:linear} reveal that adding \textit{eMargin} to InfoNCE does not lead to improved downstream classification performance. Across all three datasets—Harth, SleepEeg, and ECG—\textit{eMargin} shows consistently lower or middling scores on key evaluation metrics: accuracy, F1 score, precision, and recall. For instance, on SleepEeg, its precision (0.18) and recall (0.23) are among the lowest, despite strong clustering scores observed earlier. Similarly, on ECG, the classification performance of \textit{eMargin} lags behind TS2Vec and TNC, which both perform better on all metrics.

These results highlight a critical observation: improvements in unsupervised clustering metrics (e.g., DBI, Silhouette) do not necessarily correlate with gains in task-specific performance when representations are transferred to downstream classifiers. While \textit{eMargin} promotes tighter and more well-separated clusters in embedding space, these clusters may not align with class-relevant decision boundaries, thereby limiting their utility in supervised tasks. This discrepancy underscores the need to assess representation quality not only through unsupervised structure but also through downstream effectiveness.

\begin{figure*}[ht]
\begin{center}
\includegraphics[scale=0.155]{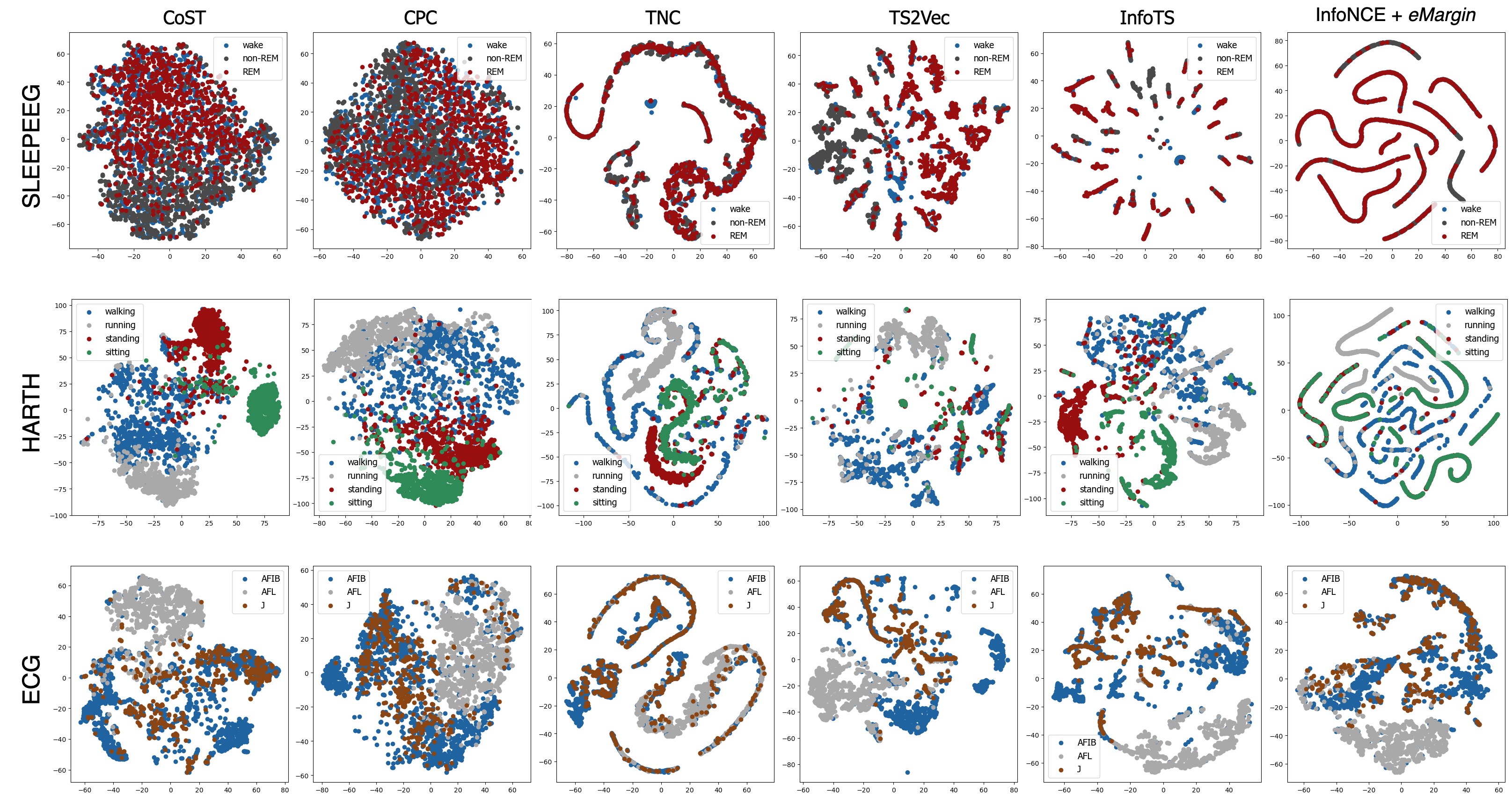}
\end{center}
\caption{t-SNE visualization of the learned embeddings on random instances on the \textsc{SleepEeg} (first row), \textsc{Harth} (second row), and \textsc{Ecg} (third row) test sets across all methods. For the \textsc{SleepEeg} each instance (data point) spans 2 seconds, while for the \textsc{Harth} and \textsc{Ecg} each instance is 1 second.}
\label{fig:tsne}

\end{figure*}

\subsection{Visualizing Representations with t-SNE}

To qualitatively assess the structure of the learned representations, we visualize the embeddings from a subset of representative models using t-SNE plots in Figure~\ref{fig:tsne}. All models were trained until convergence, and we extracted the frozen encoder representations for test samples from the three datasets: \textsc{SleepEEG}, \textsc{HARTH}, and \textsc{ECG}. The compared methods include CoST, CPC, TNC, TS2Vec, InfoTS, and our variant InfoNCE + \textit{eMargin}.

Across datasets, the t-SNE plots reveal a distinct visual pattern in the embeddings produced by \textit{eMargin}. Specifically, we observe the emergence of tight, spiral-like structures that are visually well-separated but internally continuous and elongated. These spirals are most apparent in \textsc{SleepEEG} and \textsc{HARTH}, where the class-colored traces form smooth, winding arcs. While this visual arrangement may result in low intra-cluster distance and high inter-cluster separation, thereby inflating unsupervised metrics such as Silhouette and Davies-Bouldin, it does not necessarily correspond to meaningful class separability for downstream tasks.

In contrast, other methods such as TS2Vec and TNC exhibit more discrete and semantically coherent clusters aligned with the ground-truth class labels. For instance, in the \textsc{HARTH} dataset, walking and sitting activities form distinguishable clusters under TS2Vec, whereas \textit{eMargin} produces continuous manifolds that entangle multiple activities in spiral arms.

This suggests that the apparent tightness and structure seen in \textit{eMargin}'s representations might be an artifact of its optimization objective, which promotes compactness in embedding space but neglects alignment with semantic class boundaries. As a result, although \textit{eMargin} achieves favorable scores under unsupervised clustering metrics (cf. Table~\ref{tab:cluster}), its representations generalize poorly in supervised tasks (see Table~\ref{tab:linear}).

\section{Discussion}
\label{gen_inst}


Our study investigates the impact of incorporating an adaptive margin (\textit{eMargin}) into contrastive loss for unsupervised representation learning. Our results highlight critical discrepancies between unsupervised metrics, representation structure, and downstream task performance.

One of the most striking observations comes from the t-SNE visualizations (Figure~\ref{fig:tsne}). While our proposed method, InfoNCE + \textit{eMargin}, consistently produced tight spiral-like structures in the embedding space—particularly in \textsc{SleepEEG} and \textsc{HARTH}, this visual regularity did not always translate into meaningful class separability. Despite achieving strong performance under unsupervised clustering metrics such as Silhouette and Davies-Bouldin, \textit{eMargin}'s representations did not consistently outperform other methods in supervised downstream tasks. This suggests that its optimization objective may prioritize geometric compactness and separation without capturing the underlying semantic structure necessary for generalization.

In contrast, baselines like TS2Vec and TNC, though less geometrically structured, produced clusters that better aligned with ground-truth labels, particularly in human activity recognition. These results reaffirm the importance of not over-interpreting unsupervised metrics, especially in complex, real-world data where semantic labels may follow nonlinear or overlapping distributions.

Our analysis also raises broader questions about the role of geometry in representation learning. While compact manifolds are generally desired, our findings show that the shape and structure of these manifolds—such as spirals—can introduce biases in evaluation metrics. These patterns may trick unsupervised scores into overestimating the quality of representations, particularly when the geometry favors tight local neighborhoods that do not align with global semantic separability. This reinforces that unsupervised clustering scores are, at best, (noisy) surrogate metrics and not predictors of the performance of a model in downstream tasks. Different unsupervised metrics frequently rate methods differently, complicating unsupervised model selection \cite{goswami2023}. Perhaps this is why unsupervised clustering scores are not used as an evaluation metric in many recent works on time series representation learning \cite{ts2vec,cost,tfc, simmtm, mfclr}.

In general, our findings highlight the nuanced trade-offs between compactness, structure, and semantics in learned representations. They also advocate for a more critical perspective on unsupervised evaluation protocols. Although \textit{eMargin} offers an intriguing approach to shaping latent space, future work should focus on explicitly aligning these structures with task-relevant semantics, potentially through semi-supervised objectives or better inductive biases.

\subsubsection*{Acknowledgments}

 This publication was funded by SFI NorwAI (Centre for Research-based Innovation, 309834) and the Office of Naval Research. SFI NorwAI is financially supported by its partners and the Research Council of Norway. The views expressed in this article are those of the author(s) and do not reflect the official policy or position of the U.S. Naval Academy, Department of the Navy, the Department of Defense, or the U.S. Government. We also acknowledge the useful conversations and assistance of Dr. Frank Alexander Kraemer.





\bibliography{mybibfile}

\end{document}

%% file: math_commands.tex
\usepackage{amsmath,amsfonts,bm}

\def\eqref#1{equation~\ref{#1}}

\def\1{\bm{1}}

\def\mM{{\bm{M}}}

\DeclareMathAlphabet{\mathsfit}{\encodingdefault}{\sfdefault}{m}{sl}
\SetMathAlphabet{\mathsfit}{bold}{\encodingdefault}{\sfdefault}{bx}{n}

